\documentclass[a4paper,twoside]{article}
\usepackage{epsfig}
\usepackage{subcaption}
\usepackage{calc}
\usepackage{amssymb}
\usepackage{amstext}
\usepackage{amsmath}
\usepackage{amsthm}
\usepackage{multicol}
\usepackage{pslatex}
\usepackage{placeins}
\usepackage{apalike}
\usepackage{algorithm2e}
\usepackage[bottom]{footmisc}
\usepackage{url}
\usepackage[hidelinks]{hyperref}
\usepackage{color}
\usepackage{float}
\usepackage{SCITEPRESS}   
\usepackage{orcidlink} 
\begin{document}

\title{Automatic Drywall Analysis for Progress Tracking and Quality Control in Construction}

\author{\authorname{Mariusz Trzeciakiewicz\sup{1}\orcidlink{0009-0007-6445-5759}, Aleixo Cambeiro Barreiro\sup{1}\orcidlink{0000-0002-1019-4158}, Niklas Gard\sup{1,2}\orcidlink{0000-0002-0227-2857}, Anna Hilsmann\sup{1}\orcidlink{0000-0002-2086-0951} and Peter Eisert\sup{1,2}\orcidlink{0000-0001-8378-4805}}\affiliation{\sup{1}Fraunhofer HHI, Berlin, Germany}\affiliation{\sup{2}Humboldt University of Berlin, Berlin, Germany}\email{\{mariusz.trzeciakiewicz, aleixo.cambeiro, niklas.gard, anna.hilsmann, peter.eisert\}@hhi.fraunhofer.de}}


\abstract{Digitalization in the construction industry has become essential, enabling centralized, easy access to all relevant information of a building. Automated systems can facilitate the timely and resource-efficient documentation of changes, which is crucial for key processes such as progress tracking and quality control. This paper presents a method for image-based automated drywall analysis enabling construction progress and quality assessment through on-site camera systems. Our proposed solution integrates a deep learning-based instance segmentation model to detect and classify various drywall elements with an analysis module to cluster individual wall segments, estimate camera perspective distortions, and apply the corresponding corrections. This system extracts valuable information from images, enabling more accurate progress tracking and quality assessment on construction sites. Our main contributions include a fully automated pipeline for drywall analysis, improving instance segmentation accuracy through architecture modifications and targeted data augmentation, and a novel algorithm to extract important information from the segmentation results. Our modified model, enhanced with data augmentation, achieves significantly higher accuracy compared to other architectures, offering more detailed and precise information than existing approaches. Combined with the proposed drywall analysis steps, it enables the reliable automation of construction progress and quality assessment.}

\onecolumn 
\maketitle 
\normalsize 
\setcounter{footnote}{0} 
\vfill

\section{\uppercase{Introduction}}
\label{sec:introduction}
Digitalization has become essential on modern construction sites, centralizing all the relevant building data and enabling dynamic, distributed access to vital project information. Recent efforts have aimed to enhance digitalization in construction, such as model reconstruction from scanned 2D floorplans~\cite{kalervo2019cubicasa5k,lv2021residential,barreiro2023automatic}, text analysis of construction plans~\cite{schonfelder2022deep}, and automated digitalization of fire safety information~\cite{Bayer2022}. 

Additionally, AI-based applications using digital models are emerging, demonstrating the growing demand for digitalization in modernizing the industry. Examples include leveraging digital models for autonomous robot localization~\cite{zhao2023bim}, increasing worksite automation, and localizing building interior images for automating inspection results~\cite{gard2024spvloc}.

In practice, digital frameworks support key processes throughout a construction project’s life cycle, such as real-time progress tracking and quality assurance, by maintaining up-to-date status information for building elements. This enables effective evaluation and ensures compliance with planning and quality norms over time. However, keeping information up to date is a costly task, both in terms of time and resources. This makes the automation of this field very valuable for the construction industry.

In terms of automation, image-based analysis is particularly useful, due to its potential to extract complex information from a scene without human intervention~\cite{ekanayake2021computer}. With recent advancements in deep learning techniques, such as the spread of accurate instance segmentation models, the ability to automatically analyze construction site scenes has improved significantly. 

In this paper, we focus our attention on the use of computer vision techniques for drywall analysis, with the goal of enabling automatic progress tracking and quality control. \cite{Wei2022} classifies walls into broad categories such as ``under construction" and ``finished". However, it fails to detect individual wall material types, making it impossible to estimate intermediate progress stages.
On the other hand, \cite{Ekanayake2024} distinguishes between material types but does not segment individual elements.
Another method, \cite{Pal2023}, identifies different phases of drywall construction and detects individual elements. However, the range of detected element types remains limited, and the accuracy for certain classes is relatively low.

To address the limitations of existing approaches, this paper introduces a tailored instance segmentation model for detecting and classifying individual drywall elements. To further improve model accuracy and ensure a good class coverage, particularly in scenarios with limited data, we introduce a targeted data augmentation strategy. Additionally, algorithms are introduced for drywall analysis, enabling the separation of multiple drywall segments in an image and their transformation into orthographic projections. These algorithms form the basis for a detailed analysis of the construction process, facilitating precise, segment-specific progress tracking and compliance checks with construction norms.Specifically, the contributions of this paper are:

\begin{itemize}
    \item We propose a specialized instance segmentation model for detecting and classifying individual drywall elements.
    \item We enhance the model's accuracy through a targeted data augmentation method.
    \item We develop a set of wall analysis algorithms, enabling new opportunities for automation of progress estimation and quality assurance.
\end{itemize}

\begin{figure*}[!h]
  \centering
   \includegraphics[width=\textwidth]{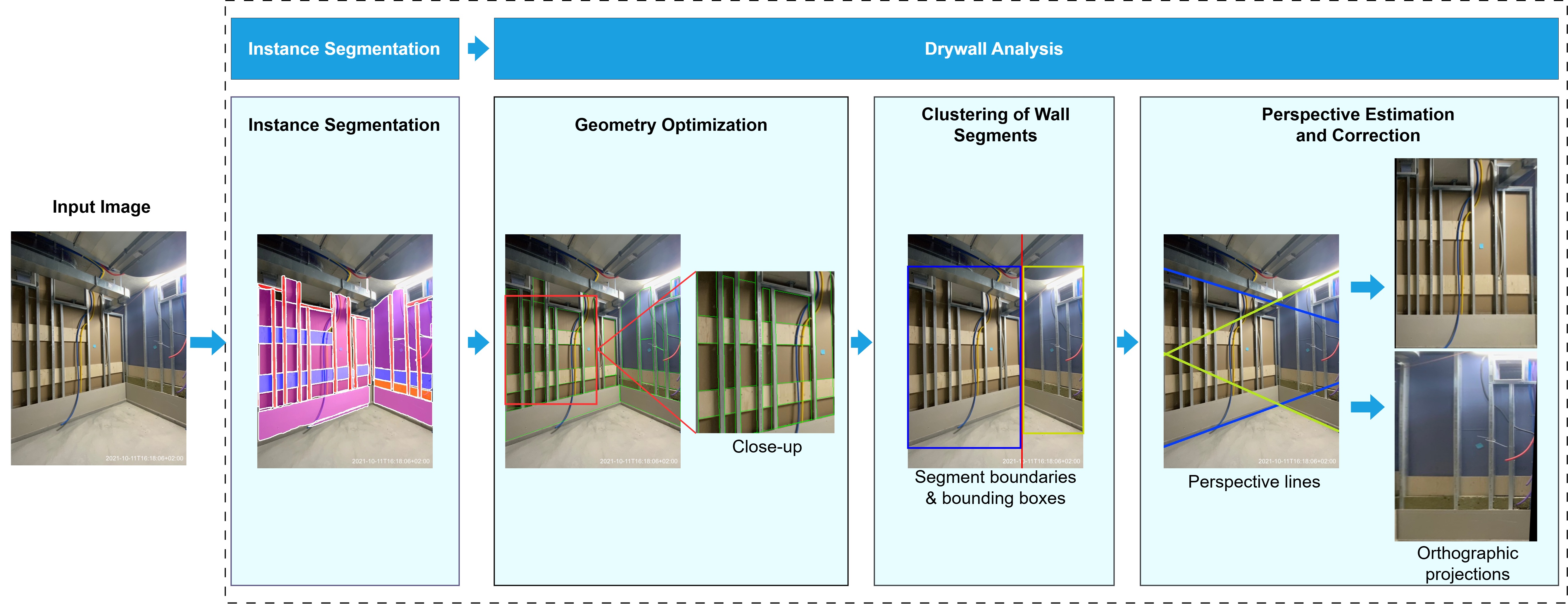}
  \caption{Instance segmentation and drywall analysis pipeline.}
  \label{fig:pipeline}
\end{figure*}

\section{Related Work}
Automated wall analysis has received some attention in the literature~\cite{Wei2022,Ekanayake2024,Chauhan2023}. In~\cite{Wei2022}, the authors employ a deep learning model for classification. However, they classify walls into general categories such as ``no construction", ``under construction", and ``finished", without detecting the types of elements used during the building process. In contrast, \cite{Ekanayake2024} presents a method that classifies different types of elements. However the authors do not distinguish between individual objects. Instead, they detect regions in the image where a certain element has been installed. While this improves wall progress analysis, separating individual elements would provide more detailed information and enable analysis of their relationships, such as checking whether metal frames are correctly spaced and parallel. Both~\cite{Shamsollahi2021} and~\cite{Ying2019} implement deep learning models for detecting and classifying individual elements on construction sites. However, \cite{Shamsollahi2021} focuses on duct detection, while~\cite{Ying2019} classifies walls, lifts, and doors.

\cite{Pal2023} highlights that most vision-based progress monitoring methods rely on a binary classification of progress as ``finished" or ``not finished", which overlooks the gradual and multi-stage nature of real-world construction. For instance, in drywall construction, it is possible to detect the metal skeleton's completion, estimate the insulation's percentage, and verify the alignment of metal frames.

A more detailed approach is proposed in \cite{Chauhan2023}, defining distinct stages like installing metal frames, drywall panels, and insulation. While this method improves stage-specific progress estimation, limited data led to lower accuracy in detecting metal frames and excluded insulation from the pipeline, hindering comprehensive performance.

Compared to~\cite{Wei2022}, which focuses on general wall classifications, we detect and classify key drywall elements for progress and quality estimation, including ``drywall panels", ``metal frames", ``wood panels", and ``insulation materials". In contrast to~\cite{Ekanayake2024}, our method separates each individual drywall element, enabling a detailed analysis of how their relationships. We also introduce a data augmentation strategy to address dataset limitations. In comparison with~\cite{Chauhan2023}, our approach effectively detects all key types of drywall elements. Furthermore, unlike other approaches, we cluster and correct perspective distortions in drywall segments, enabling detailed progress estimation and quality assurance in drywall construction.

\section{\uppercase{Methodology Overview}}
\label{sec:methodology_overview}

Figure~\ref{fig:pipeline} illustrates the proposed pipeline, which consists of two main components: instance segmentation and drywall analysis. The instance segmentation component, described in detail in Section~\ref{sec:component_detection}, uses a modified Mask R-CNN model (Section~\ref{sec:model_architecture}) to detect individual drywall components, enhanced by task-specific data augmentation techniques (Section~\ref{sec:data_agumentation}). In the drywall analysis component, detailed in Section~\ref{sec:drywall_analysis}, we process the output of the instance segmentation model. First, we simplify and optimize the detected geometries based on prior knowledge (Section~\ref{sec:geometry_optimization}). Then, using these optimized geometries and the orientation of their edges, we separate drywall segments when multiple are present in the image (Section~\ref{sec:segment_clustering}). Finally, based on the refined geometries, we estimate perspective distortions caused by varying drywall angles relative to the camera and correct these distortions to obtain orthographic projections (see Section~\ref{sec:pers_estimation_correction}).

\section{Dataset}
\label{sec:dataset}
We use a dataset of 176 drywall images, divided into two subsets: 140 images for training and 36 for testing. Each image, with resolutions of 600$\times$800, 800$\times$450, or 800$\times$600 pixels, is manually annotated with class labels, segmentation masks, and bounding boxes for each element of four essential classes: wood panel, insulation, drywall panel, and metal frame. These classes represent the main elements involved in the construction of a drywall, and are crucial for accurate progress and quality estimation.

These images capture drywalls at various stages of construction, some of them containing multiple wall segments. The majority depict walls in an intermediate phase, with interior layers exposed, often showing elements from each class. A few images show only the metal frame skeleton, while others display completed, plastered walls ready for painting.

The main challenges of the dataset, which motivated the development of the proposed data augmentation technique (Section~\ref{sec:data_agumentation}), are its limited size and the variability of materials within the same class.

In addition to this dataset, we obtained a separate video recording documenting the construction of a drywall structure for a new room within a building. Selected frames from this recording were used to assess the generalization capabilities of our drywall analysis pipeline on scenes outside the training dataset. This approach allows us to evaluate the practical viability of our method for application in real-world construction projects.

\section{\uppercase{Component Detection}}
\label{sec:component_detection}
We propose a modified deep learning architecture for detecting drywall elements in camera images, incorporating task-specific enhancements to improve performance. Additionally, we introduce a drywall-specific data augmentation technique that increases dataset variability, addressing challenges posed by limited training data and enhancing model accuracy.

Accurate drywall analysis depends on extracting high-quality, detailed information from images, enabling precise results and enriching digital models with valuable metadata for tracking and quality control. To achieve this, we focus on detecting and classifying individual drywall elements with high spatial accuracy. Instance segmentation, chosen for its pixel-wise masks, provides the precision needed to meet the stringent requirements of automated drywall analysis, surpassing simpler bounding-box detection methods.
\subsection{Model Architecture}
\label{sec:model_architecture}

\begin{figure}[!htb]
  \centering
   \includegraphics[width=\columnwidth]{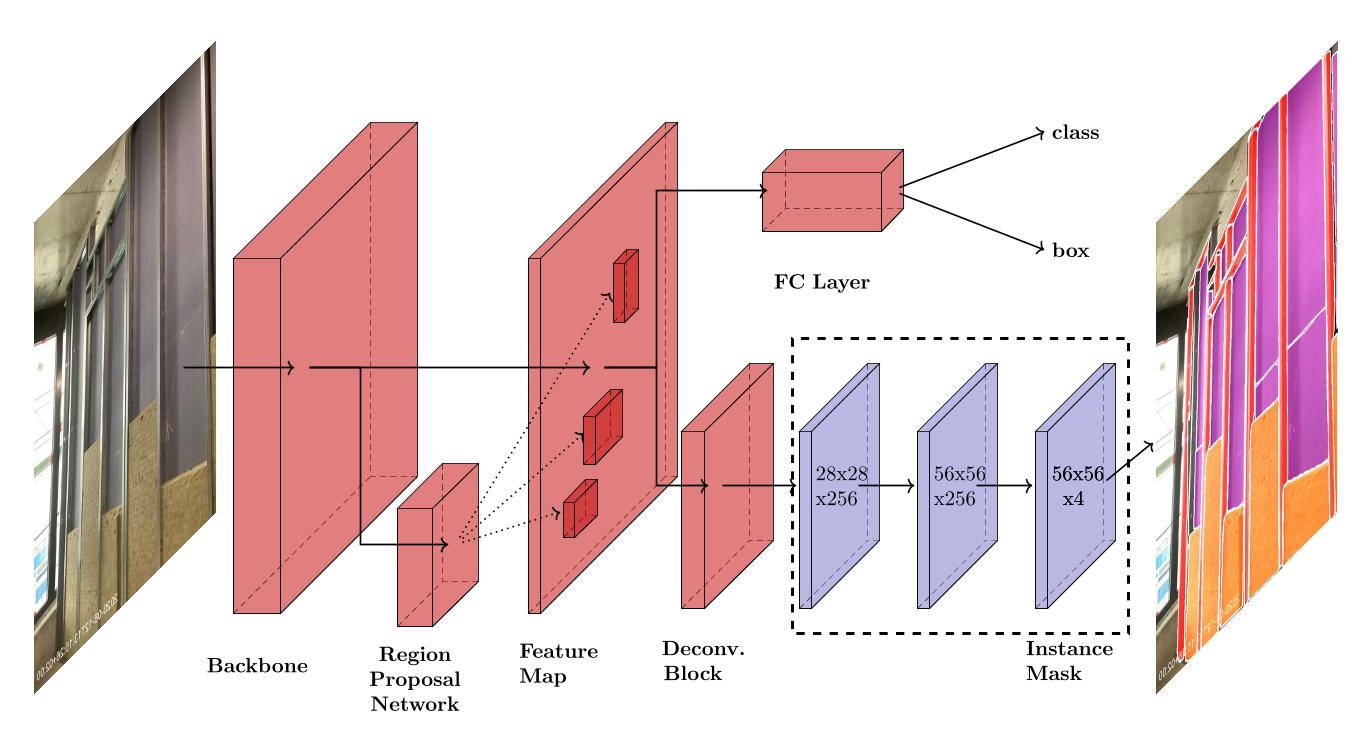}
  \caption{Proposed modification of the Mask R-CNN architecture. We replace the backbone used in the original paper with the ConvNeXt V2 model, introduce additional ratios for computing anchor boxes, and add a block of deconvolutional layers at the end of the instance mask branch to increase the output mask resolution.}
  \label{fig:mask_rcnn_model}
\end{figure}

For instance segmentation, we employ a modified version of Mask R-CNN~\cite{8237584}, an architecture whose variants with different backbones achieve state-of-the-art results~\cite{Dalva2023} on widely recognized benchmark datasets such as COCO~\cite{lin_microsoft_2014}. This choice was motivated by the model's strong performance and adaptability across various segmentation tasks.

The modified architecture is illustrated in Figure~\ref{fig:mask_rcnn_model}. The first modification involves replacing the original backbone model. After evaluating several options, ConvNeXt V2~\cite{Woo2023} yielded the best results on our dataset, leading us to select it as the backbone for our model.

The second proposed modification accounts for the unique characteristics of the elements in the dataset and the way Mask R-CNN calculates region proposals. To detect objects in images, Mask R-CNN contains a block of convolutional layers called the Region Proposal Network (RPN) that identifies potential object locations by generating a set of anchor boxes, distributed over the entire image, based on a list of predefined ratios. Each of these ratios is responsible for detecting objects of different shapes. The RPN is then trained to predict whether each anchor box corresponds to an object or background, and to adjust the anchor box coordinates to match the object's size.

In our dataset, most objects share similar, typically rectangular shapes, aligning well with the anchor box shapes in the original model. However, metal frames are usually long and narrow, making them less suited to the original anchor box configurations and harder for the model to detect accurately. To address this, we introduce additional ratios during the anchor boxes generation step allowing the model to better capture the distinctive shape of metal frames.

The final proposed modification to the model addresses the size of the predicted instance masks. The original instance segmentation branch of Mask R-CNN produces low-resolution binary masks with a fixed size of 28$\times$28 pixels. These masks are then upscaled to match the dimensions of the detected object in the original image. This incurs a loss of spatial resolution that often introduces artifacts, especially for larger elements, causing distortions in object outlines. To reduce these artifacts, we add a set of deconvolution layers at the end of the instance segmentation branch, increasing the resolution of the masks to 56$\times$56 pixels before the upscaling step. This modification results in smoother and more accurate object borders.

\subsection{Data Augmentation}
\label{sec:data_agumentation}

\begin{figure}[!t]
  \centering
   \includegraphics[width=\columnwidth]{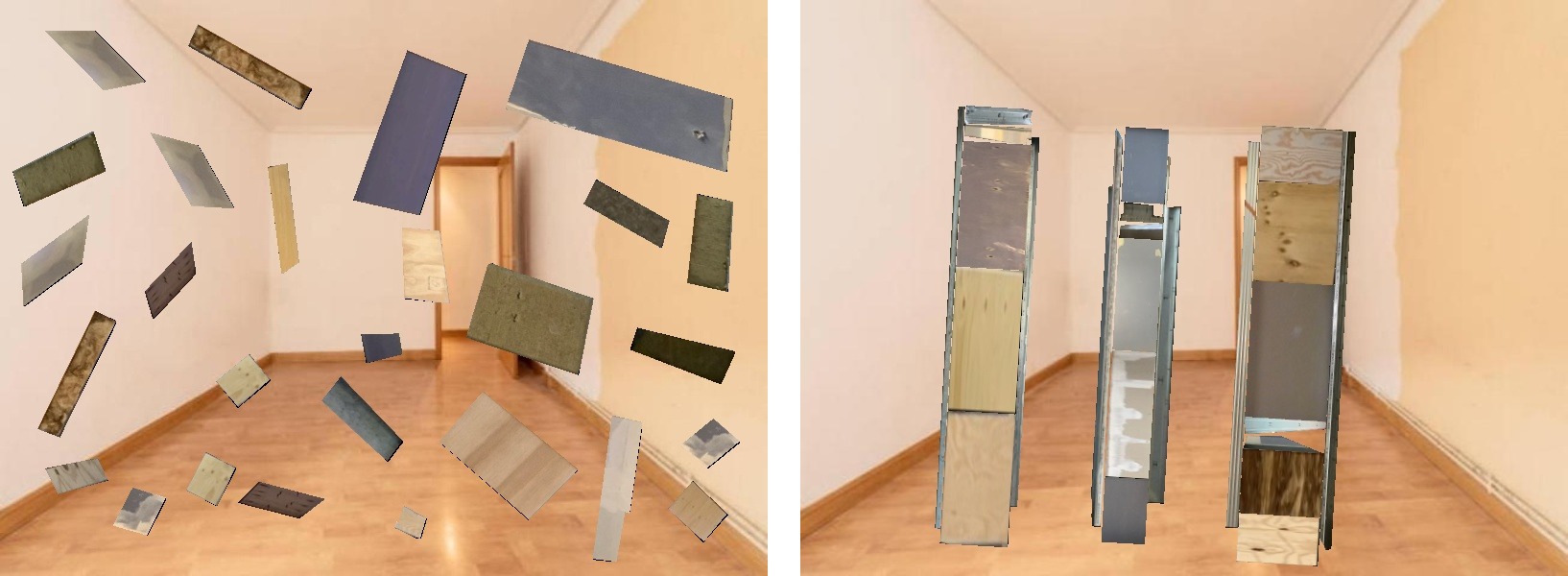}
  \caption{Example of synthetic images. The left image shows augmentation with randomized placement of cropped elements, while the right image demonstrates structured placement.}
  \label{fig:data_augmentation}
\end{figure}

The number of images in our dataset is limited, with 176 samples to train and validate our model. For comparison, in~\cite{8237584}, the authors trained their Mask R-CNN model on the COCO dataset, which contains 330,000 images. This scarcity of training data poses a challenge in achieving a model with good generalization capabilities.

A common approach to address training data scarcity, particularly when acquiring additional samples is unfeasible or costly, is to employ data augmentation techniques. In addition to basic methods, such as random rotation, flipping, or scaling of training images, we here adapt a more sophisticated technique introduced in~\cite{Barreiro2022}. To generate new training samples, labeled samples are selected from the training set that constitute ``good examples", i.e. with unambiguous labels and covering sufficient appearance variations. These selected examples are then used to generate synthetic images. To this end, randomized transformations such as rotation, translation and scaling are applied to the examples, which are then pasted onto backgrounds consisting of a collage of random COCO images, increasing the variety of a scarce dataset.

Drywall elements are, however, usually arranged in a specific, structured layout, providing additional contextual information that can be leveraged to improve training. Insulation is usually placed between parallel, evenly spaced metal frames, with wall and wood panels attached to them. Randomly positioning the objects in a synthetic image disrupts this specific structure of a drywall, reducing realism. For this reason, we modify the aforementioned data augmentation method to generate synthetic images that more accurately reflect the organized configuration of drywall construction.

Following the approach in~\cite{Barreiro2022}, we first select suitable elements using the same criteria as the original method. Instead of randomly pasting the selected objects onto new images, we insert them in a more structured way. We start by sampling long metal frames, positioning them vertically and parallel at random intervals. The space between adjacent frames is divided into rectangular slots, filled with randomly selected, scaled examples from various classes. Occasionally, spaces are left empty to increase dataset variety. Finally, the columns are rotated randomly around the image center. Instead of COCO,  we use a dataset that contains images of empty rooms, called room-interior~\cite{room-interior_dataset}. This allows the model to distinguish drywall elements from common background objects found in indoor environments, such as painted walls, ceilings and floors. An example of a synthetic image generated using this method is shown in Figure~\ref{fig:data_augmentation}.

\section{\uppercase{Drywall Analysis}}
\label{sec:drywall_analysis}

Once the construction elements are localized in an image, we can analyze them to extract information relevant to the digital model, enabling tasks such as progress evaluation or quality control based on predefined standards.

Some images include multiple wall segments, which presents a unique challenge. Ideally, the information for each individual segment should be stored separately to increase the level of detail in the digital model, since they may be at different construction stages. Therefore, we must first identify the elements belonging to each segment. Additionally, each segment is captured from a different angle, resulting in varying perspectives. By isolating segments and correcting their perspective distortions, we generate an orthographic projection of each, enabling more complex analysis. This allows us to detect structural issues that would be difficult to assess in perspective-distorted images, facilitating automated quality control by verifying, for example, that metal frames are parallel and installed at the correct intervals.

It is trivial to differentiate wall segments that are physically separated from one another. Walls forming a corner can be distinguished by their perspective distortions. In an upright camera perspective view of a wall segment, all associated horizontal lines converge at a common vanishing point, which varies for each segment if they are at different angles. The horizontal sides of the detected elements provide an estimation of the horizontal lines in the wall. Therefore, we refine these edges and use them to separate the wall segments. 

\subsection{Geometry Optimization}
\label{sec:geometry_optimization}

As mentioned, we utilize the horizontal sides of detected elements to estimate vanishing points in order to separate drywall segments. However, these detected elements often exhibit irregularities along their borders. Therefore, we have implemented an algorithm that simplifies and optimizes these shapes.

The algorithm takes the polygons representing the outlines of the detected elements and finds 4 corner candidates for each by iteratively adding contiguous vertices to a set until they cannot accurately be fit by a straight line. It then uses RANSAC~\cite{fischler1981random} to fit lines through the vertices between the candidates, minimizing the impact of outliers. Thus, we obtain accurate 4-sided polygons representing the detected elements.

To further increase accuracy, we leverage prior knowledge of drywall structure. Specifically, neighboring elements share sides and are often arranged in rows or columns with edges aligned along the same horizontal or vertical line. We group these aligned edges and, for each group, we collect their endpoints and apply RANSAC to fit a line through them. The original edges are then replaced with the newly calculated lines. Finally, we calculate the intersection points of updated sides with the other sides to form the refined polygons.

\subsection{Clustering of Wall Segments}
\label{sec:segment_clustering}

In this section, we describe our method for estimating vanishing points for individual drywall segments using the horizontal edges of optimized polygons. By identifying these points, we can assign each element to its corresponding drywall segment, thereby clustering distinct drywall segments within a single image.

Depending on their angles with respect to the camera position, wall segments will have different vanishing points where the horizontal lines converge. Identifying these points enables us to determine the edges that align with them and correctly associate polygons with their respective wall segments. To achieve this, we divide the elements in the image into vertical columns. Within each column, we collect the edges that lie fully or partially within it and calculate their pairwise intersections, which allows us to determine a vanishing point for each column sequentially.

If neighboring columns share the same vanishing point, they belong to the same segment. Conversely, if two adjacent columns have different vanishing points, the boundary between wall segments lies between them. In cases where two segments converge within a single column, the intersection points of the edges within it are scattered across a larger area instead of being concentrated in one location. Therefore, if we detect such a case, we divide the column into smaller parts and repeat the vanishing point calculation for each.

\subsection{Perspective Correction}
\label{sec:pers_estimation_correction}

As the last step of the proposed drywall analysis, we obtain orthographic projections of each drywall segment, opening up new possibilities for automating progress monitoring and quality control. To achieve this, we first estimate the perspective of each individual wall segment, then correct it to remove perspective distortions.

Most drywall elements and the walls themselves are rectangular and approximately axis-aligned within the same plane. As a result, the perspective distortion affecting the entire wall segment, as well as the transformation necessary to correct it, is approximately the same for the individual elements that are part of it. Therefore, a simple approach to obtaining the segment's orthographic projection is to obtain this of one of its constituting elements, which can be done by calculating the homography that maps it to a rectangular, axis-aligned box. This method is, however, quite sensitive to small inaccuracies and limits imposed by finite resolution, especially so when the chosen element does not cover a big portion of the wall’s surface. In order to tackle this, it is possible to combine the transformations calculated for each of the elements belonging to the wall segment. To this end, we find the wall’s corners within the orthographic projection calculated for each element and project them back to the original image. Then, we use RANSAC to calculate the average position for each corner minimizing the effect of outliers. Finding the corners of the wall in an orthographic projection is a straightforward process: the minimum and maximum values along the X and Y axis of its constituting elements correspond to the limits of its bounding-box, which is axis-aligned. Once the four corners of the wall are calculated for the original image, a homography transformation can be calculated to obtain its orthographic projection.

\section{\uppercase{Results}}
\label{sec:results}

In this section, we present the results of our method, structured into two subsections: one in which we discuss the performance of the instance segmentation module, and one in which we discuss this of the wall analysis module. In the former, the effectiveness of the modifications made to the Mask R-CNN model and the impact of data augmentation on its accuracy will be reviewed. In the latter, we demonstrate the outcomes of each post-processing step applied during the wall analysis.

\subsection{Instance Segmentation}
In this subsection, a comparison of performance between different backbone choices for the Mask R-CNN model will be briefly discussed. Based on this, the one offering best results (ConvNeXt V2) is chosen as a baseline, against which we evaluate the impact of each proposed modification to the network architecture, as well as the data augmentation methods. This evaluation is conducted at both the quantitative and qualitative levels.

\subsubsection{Quantitative Evaluation}

For quantitative evaluation, we use mean average precision (mAP), a common metric for instance segmentation models. Specifically, we apply mAP@0.5:0.95, which computes mAP across intersection over union (IoU) thresholds from 0.5 to 0.95 in increments of 0.05 and then averages the results. The IoU threshold is used to decide whether a detection is true positive (TP) of false positive (FP). If IoU between the ground truth and detected object is smaller than the threshold, it is considered as FP, otherwise TP. We provide mAP for both bounding boxes (bbox mAP) and instance masks (mask mAP).

Table~\ref{tab:maskrcnn_backbones} provides an overview of the accuracy of Mask R-CNN with different backbones on our dataset. All backbones were pretrained on the COCO dataset. Among the various backbone architectures, ConvNeXt V2 clearly yields the best results, which is why we chose it as a baseline for our method.

\begin{table}[h]
\caption{Accuracy results for Mask R-CNN with different backbones on our dataset.}\label{tab:maskrcnn_backbones} \centering
\begin{tabular}{|c|c|c|}
  \hline
  Backbone & bbox mAP  & mask mAP \\
  \hline
  FPN & 0.47  & 0.45\\
  \hline
  Swin & 0.40  & 0.39\\
  \hline
  ResNest & 0.40  & 0.40\\
  \hline
  \textbf{ConvNeXt V2} & \textbf{0.55}  & \textbf{0.52}\\
  \hline
\end{tabular}
\end{table}

Table~\ref{tab:maskrcnn_modifications} shows how the proposed model modifications and data augmentation techniques influenced accuracy compared to the baseline. All models were trained for 60 epochs with early stopping and ConvNeXt V2 backbone. The initial learning rate was set to 0.02 and decreased by a factor of 10 at the 50th and 55th epochs. We used a weight decay of 0.001 and a momentum of 0.9.

Adding aspect ratios for anchor boxes to improve narrow frame detection, as well as increasing the instance mask output size, led to a slight improvement in the accuracy of both bounding box and instance mask detections. However, the most significant improvement was achieved by incorporating synthetically created images to the training as data augmentation. The base synthetic generation method, as described in~\cite{Barreiro2022}, improved bbox mAP by approximately 5\% and mask mAP by around 4\% on top of the improvements achieved by the modified baseline with only basic augmentation techniques. Further accuracy gains were obtained by using synthetic backgrounds that more closely resemble the scenes from the original images, and replacing the random placement of cropped elements with a structured arrangement that imitates actual drywall patterns. These enhancements, combining realistic synthetic backgrounds with a more natural arrangement of elements, enabled the model to produce more accurate instance masks and achieve increase recall.

\begin{table}[h]
\caption{Evaluation results of Mask R-CNN models with various modifications and data augmentation strategies. Here, B represents the baseline model; EB extends the baseline by adding additional anchor ratios and increasing output mask sizes through deconvolution; CB incorporates data augmentation by blending objects onto COCO dataset backgrounds; RB employs data augmentation with room-interior backgrounds; RP uses data augmentation with randomized placement, and SP data augmentation with structured placement.}\label{tab:maskrcnn_modifications} \centering
\begin{tabular}{|c|c|c|}
  \hline
  Model & bbox mAP  & mask mAP \\
  \hline
  B &  0.55  & 0.52 \\
  \hline
  EB & 0.57  & 0.54\\
  \hline
  EB + CB + RP & 0.62  & 0.58\\
  \hline
  EB + CB + SP & 0.63  & 0.59\\
  \hline
  EB + RB + SP & \textbf{0.64}  & \textbf{0.60}\\
  \hline
\end{tabular}
\end{table}

\subsubsection{Qualitative Evaluation}

\begin{figure}[!t]
  \centering
   \includegraphics[width=\columnwidth]{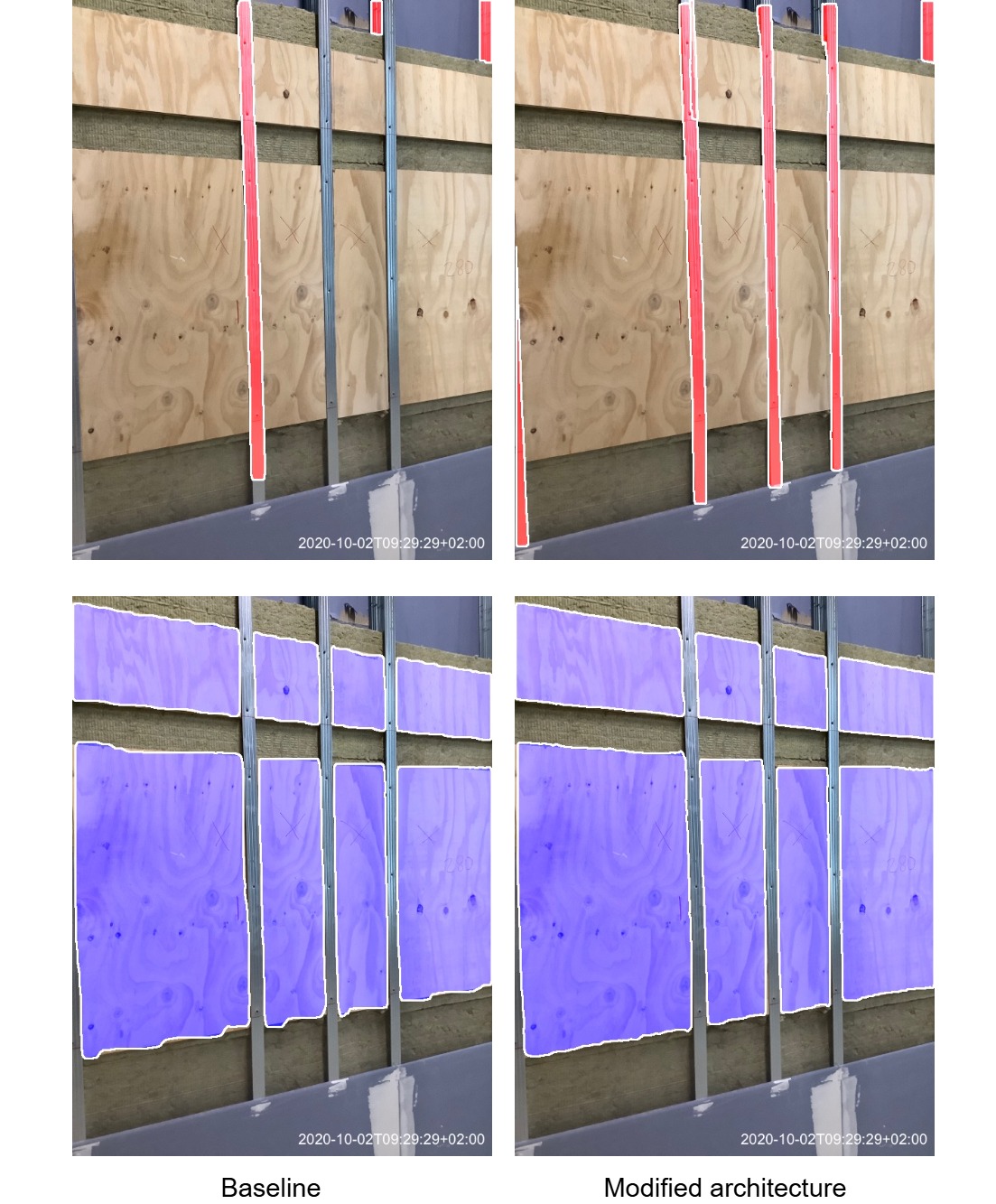}
  \caption{Visual comparison of detected metal frames (first row, in red) and wood panels (second row, in blue) by the baseline model and the modified architecture.}
  \label{fig:model_modifications}
\end{figure}

\begin{figure}[!t]
  \centering
   \includegraphics[width=\columnwidth]{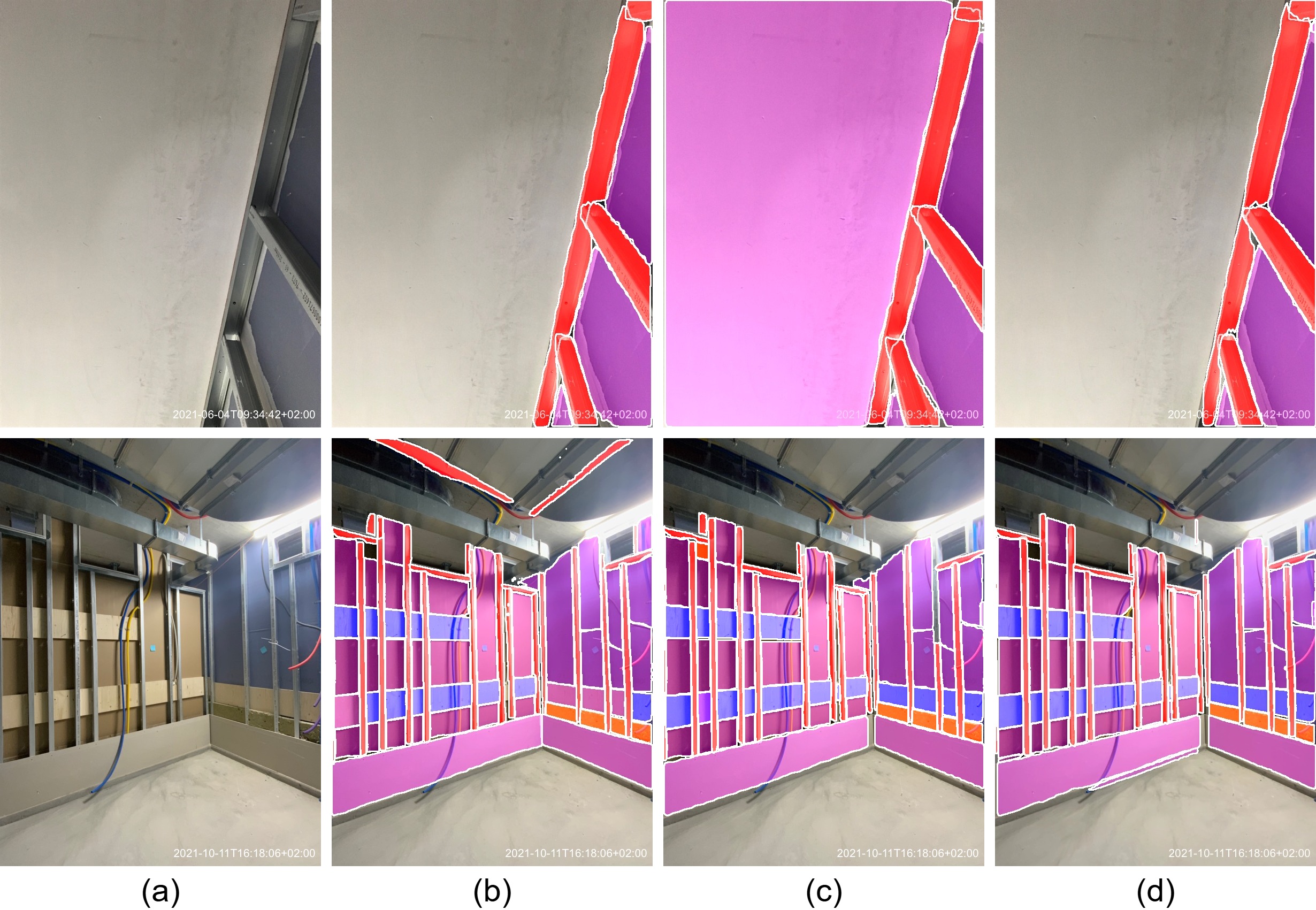}
  \caption{Instance segmentation results using different data augmentation techniques. Column \textbf{(a)} shows the original image, \textbf{(b)} the baseline output, \textbf{(c)} the baseline with data augmentation using COCO backgrounds and random placement, and \textbf{(d)} the baseline with data augmentation using room-interior backgrounds and structured placement. Detected wall panels are shown in pink, wood panels in blue, insulation in orange, and metal frames in red.} 
  \label{fig:models_comparision}
\end{figure}

The impact of the proposed architecture modifications is reflected in the quantitative results. However, this impact is even more evident when we compare the instance masks produced by the baseline model and its improved version. A visual comparison is shown in Figure~\ref{fig:model_modifications}. The addition of extra aspect ratios to generate anchor boxes more suitable for narrow metal frames was essential in improving the recall for this class. On the other hand, we can see that for the baseline model with masks at 28$\times$28 pixels, visible artifacts appear along the borders of detected elements. Increasing the instance mask output size to 56$\times$56 pixels through deconvolution resolved this issue, as the detected polygons no longer exhibit the ``staircase" effect seen in the figure. This is particularly relevant for a clear element edge determination, which is vital to the accuracy of our wall analysis algorithm.

Figure~\ref{fig:models_comparision} illustrates the influence of data augmentation on instance detection using the baseline model. Overall, the baseline model delivers strong detection results, accurately identifying most elements within the drywall with only a few misclassifications. Moreover, it can even distinguish between adjacent wall panels with minimal, barely visible borders.

The model trained with data augmentation, nevertheless, detects fewer false positives and shows fewer mismatched class labels compared to the model trained without augmentation. However, the augmentation method that creates new images by randomly placing elements onto COCO backgrounds disrupts the characteristic structure of drywalls. As a result, some elements that clearly do not belong to any wall segment (i.e.\ those outside the segment borders) may be falsely identified as drywall elements. This happens because the model learns that some elements can appear randomly distributed across the entire image. Our modified augmentation approach addresses this issue by teaching the model that drywall elements are typically surrounded by other components following a certain structure, e.g., insulation is always placed between metal frames.

\subsection{Drywall Analysis}
For the evaluation of the proposed drywall analysis process, we focus on the qualitative results that allow us to estimate the viability of this method for practical applications, such as progress tracking and quality control. To this end, we include visualizations of the results of the most important steps and provide a discussion of the findings.

\begin{figure}[!htb]
  \centering
   \includegraphics[width=\columnwidth]{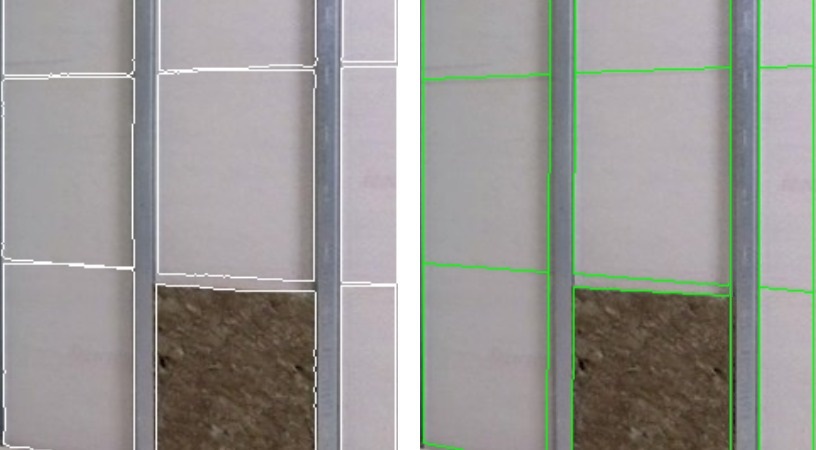}
  \caption{Results of the group refinement of simplified polygons. Borders of the detected wall panels are shown in white, and optimized polygons in green.}
  \label{fig:res_polygon_optimization}
\end{figure}

As shown in Figure~\ref{fig:res_polygon_optimization}, our polygon refinement technique significantly enhances the accuracy and consistency of the outlines of detected elements. The horizontal edges of these polygons are precisely aligned with each other and with the intersecting horizontal lines that converge at the corresponding vanishing point. This alignment greatly improves the estimation of vanishing points in the subsequent steps.

\begin{figure}[!t]
  \centering
   \includegraphics[width=\columnwidth]{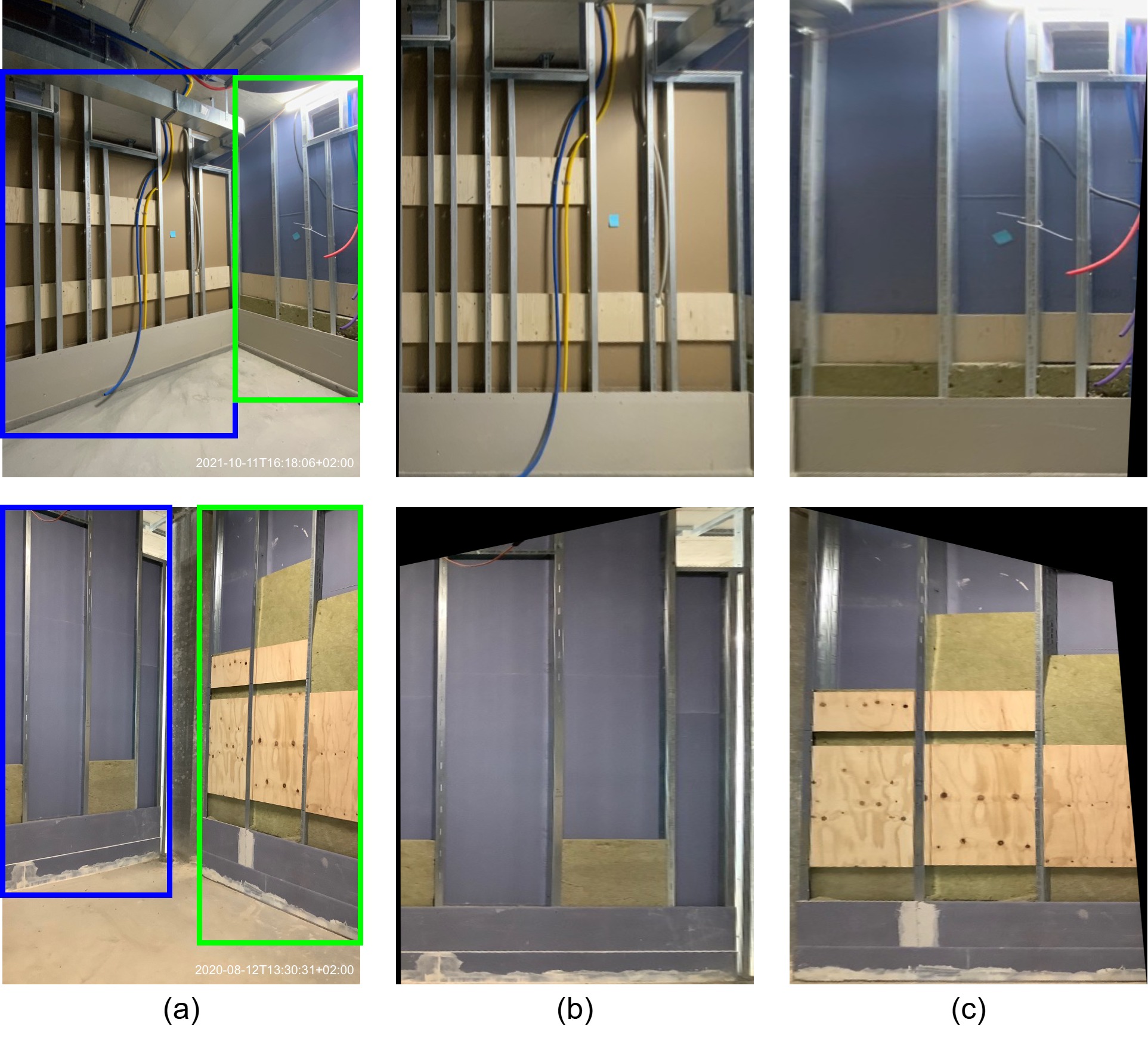}
  \caption{Visualization of our drywall segment clustering and perspective correction results. Column \textbf{(a)} displays the original images with two bounding boxes in blue and green indicating the different wall segments. Column \textbf{(b)} shows the undistorted view of left segments, while column \textbf{(c)} shows this of right segments.}
  \label{fig:wall_analysis}
\end{figure}

\begin{figure*}[!t]
  \centering
   \includegraphics[width=0.9\textwidth]{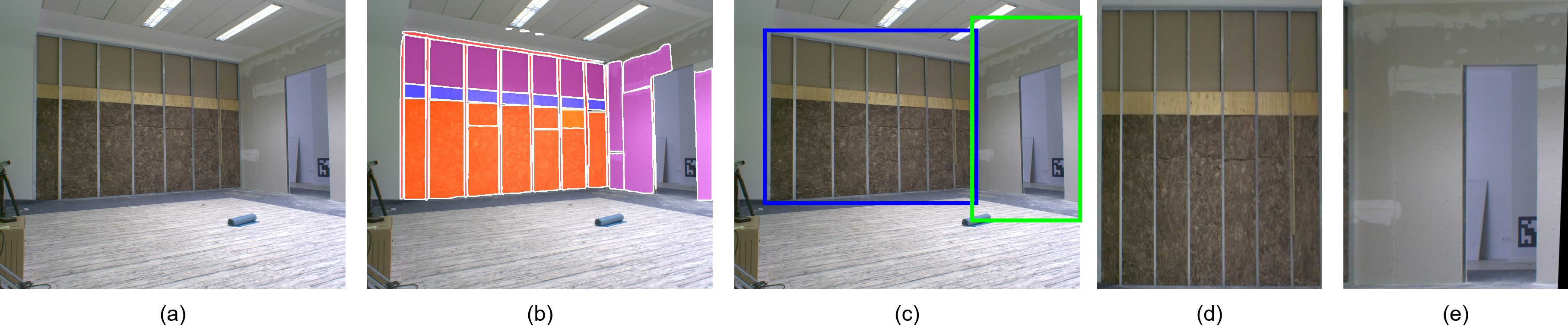}
  \caption{Visualization of the complete instance segmentation and wall analysis pipeline on new data. In this example: \textbf{(a)} shows the original image, \textbf{(b)} displays the instance segmentation output, \textbf{(c)} illustrates the bounding boxes of the two wall segments, and \textbf{(d)} and \textbf{(e)} present the segments with corrected perspective distortions.}
  \label{fig:own_image_pipeline}
\end{figure*}

With accurately estimated polygons, we can successfully apply our drywall segment clustering and perspective correction algorithms. Figure~\ref{fig:wall_analysis} shows examples from our dataset with two visible wall segments. In both cases, we successfully identified the borders between segments, estimated perspective distortions, and corrected them. The resulting projections show elements that are parallel to each other, creating new opportunities for analyzing individual drywall elements and their relationships, which is crucial for progress and quality analysis.

\section{\uppercase{Discussion}}
\label{sec:discussion}
 Our model demonstrates robust instance segmentation performance on our dataset. Additionally, the proposed wall analysis method successfully isolates individual drywall segments and corrects their perspective distortions. However, the limited number of testing images in the dataset poses challenges in fully assessing the system’s generalization capabilities. To address this, we captured the construction process of a drywall structure to create a new room by installing on-site cameras at fixed locations. We then applied the entire pipeline to these new images representing different stages of construction captured from different points of view. Figure~\ref{fig:own_image_pipeline} offers a visualization of the results for one of these images.
 
 We observe that our model not only successfully detects and classifies most drywall elements but also effectively uses this information to separate the two wall segments present in the image and obtain their orthographic projections. This demonstrates that our proposed instance segmentation model and analysis algorithm generalize well to data the model has never encountered before, and it would be viable to use our system in real construction projects to automate data collection.

 Compared to \cite{Wei2022,Ekanayake2024}, our segmentation model not only detects more types of drywall elements but also distinguishes individual elements and estimates their precise location in the image. Combined with our drywall analysis module, it enables a more comprehensive progress estimation than simply classifying walls as ``finished" or ``not finished." Compared to~\cite{Chauhan2023}, our method successfully detects metal frames and includes insulation detection, offering more detailed information on the drywall construction process.
 
Figure~\ref{fig:own_image_pipeline} illustrates the results of our pipeline applied to a frame of our recorded drywall construction process. Utilizing a calibrated camera with known positioning, the extracted information is directly mapped to a corresponding wall in a digital model, automating status updates.  Furthermore, the pipeline effectively detects individual elements within each wall segment with high accuracy. This data, such as the surface area covered by insulation, can then be compared against digital model plans to evaluate construction progress and ensure compliance with project specifications.

\begin{figure}[!htb]
  \centering
   \includegraphics[width=.98\columnwidth]{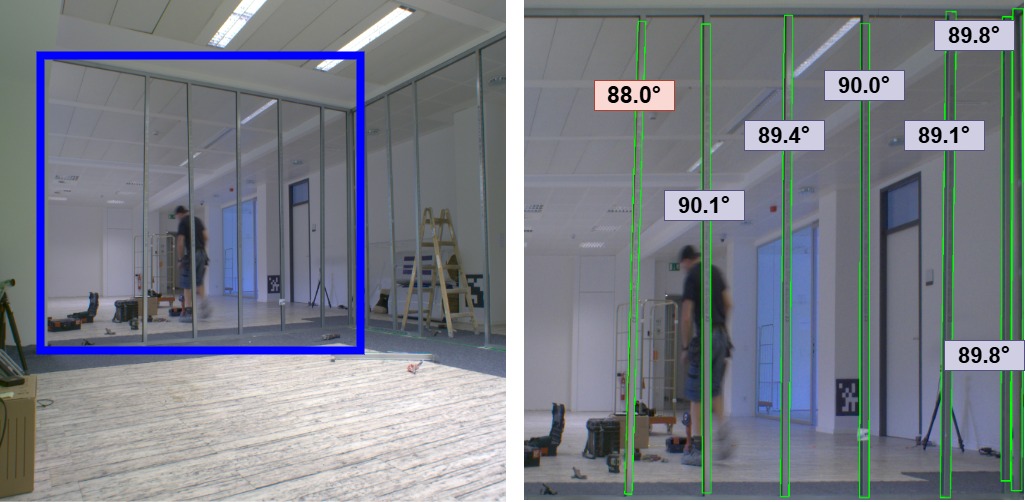}
  \caption{An example of a quality assurance use case employing the proposed method. The left image shows the segment under analysis highlighted in blue. The right image illustrates the detected vertical metal frames in green, along with their angles (in degrees).}
  \label{fig:quality_assurance}
\end{figure}

To demonstrate a practical application of our wall analysis method for quality assessment, we analyzed an image showing only metal frames, as seen in Figure~\ref{fig:quality_assurance}. We generated an orthographic view of a specific wall segment from the image. Using this view, we estimated the orientation of the metal frames. This analysis revealed that one of the frames exhibited a tilt exceeding 1$^\circ$ (highlighted in red). Based on this information a warning can be issued, enabling corrective action to be taken before further progress is made in the wall construction process.

\section{\uppercase{Conclusions}}
In this paper, we present an automated pipeline for drywall analysis using on-site cameras, combining a deep learning instance segmentation module with an analysis module for extracting valuable information to create rich digital models for progress tracking and quality control. We enhanced a standard neural network architecture and developed targeted data augmentation to improve segmentation performance with limited training data. Additionally, we introduced a novel algorithm to extract useful information from the segmentation results, to support automatic progress tracking and quality control, as shown in real construction-site applications. Our method provides more detailed information than existing state-of-the-art approaches, and we demonstrate generalization to scenes outside the training dataset, showcasing its potential to boost productivity, planning accuracy, and regulatory compliance in the construction industry.

\section*{\uppercase{Acknowledgements}}
This work has partly been funded by the German Federal Ministry for
Digital and Transport (project EConoM under grant
number 19OI22009C). We thank HOCHTIEF ViCon GmbH for providing training images of construction sites.

\bibliographystyle{apalike}
{\small
\bibliography{example}}

\end{document}